# Cheating for Problem Solving:
# A Genetic Algorithm with Social Interactions

Rafael Lahoz-Beltra
Department of Applied Mathematics
Faculty of Biological Sciences
Complutense University of Madrid
Madrid 28040, Spain

lahozraf@bio.ucm.es

Gabriela Ochoa
Automated Scheduling, Optimisation
and Planning (ASAP) Group
School of Computer Science
University of Nottingham, UK

gxo@cs.nott.ac.uk

Uwe Aickelin
Intelligent Modelling & Analysis
Research Group (IMA)
School of Computer Science
University of Nottingham, UK

uwe.aickelin@nottingham.ac.uk

## ABSTRACT
We propose a variation of the standard genetic algorithm that incorporates social interaction between the individuals in the population. Our goal is to understand the evolutionary role of social systems and its possible application as a non-genetic new step in evolutionary algorithms. In biological populations, i.e. animals, even human beings and microorganisms, social interactions often affect the fitness of individuals. It is conceivable that the perturbation of the fitness via social interactions is an evolutionary strategy to avoid trapping into local optimum, thus avoiding a fast convergence of the population. We model the social interactions according to Game Theory. The population is, therefore, composed by cooperator and defector individuals whose interactions produce payoffs according to well known game models (prisoner's dilemma, chicken game, and others). Our results on Knapsack problems show, for some game models, a significant performance improvement as compared to a standard genetic algorithm.

## Categories and Subject Descriptors
I.2.8 [Artificial Intelligence]: Problem Solving, Control Methods, and Search – *Heuristic methods*

## General Terms
Algorithms, Design.

## Keywords
Genetic algorithms, social interaction, game theory, knapsack problems.

## 1. INTRODUCTION
Genetic Algorithms (GAs) are designed to search for near-optimal solutions in search spaces with multiple local optima. The GA population endows the algorithm with noise-resistant properties. According to [1], when the fitness values are modified by the addition of a noise term N(0, $\sigma^2$), the algorithm's performance decreases in proportion to the value of $\sigma$. To the best of our knowledge, any study so far has addressed the issue of how to improve the performance of GAs or other evolutionary methods based on a 'perturbation' of the fitness values.

This paper proposes a new approach inspired by the behavior of individuals in social systems. Using the well-known knapsack problem, we show how the inclusion of 'social interactions' into the GA cycle, significantly improves the algorithm's performance. Our approach is loosely related to co-evolutionary approaches in that the fitness of an individual depends on its relationship to other members of the population [2, 3]. However, it differs from them in that the social interaction is not the main mechanism for calculating the fitness of co-evolving (competing or cooperating) species. Instead, in our approach the social interaction is an additional step that slightly alters the individuals' genetic fitness values.

It is worth pointing out that we are not interested in competing with state-of-the-art heuristic approaches for the Knapsack problem [4, 5]. Our interest lies instead in exploring how social interactions modeled via Game Theory (GT) could improve the GA performance. We aim to understand the evolutionary role of social systems and its possible application as a non-genetic new step in evolutionary algorithms. The main inspiration for our approach is the following observation: in biological populations, i.e. animals, even human beings and microorganisms, social interactions often affect the fitness of individuals [6]. It is conceivable that the perturbation of the fitness via social interactions is an evolutionary strategy to avoid trapping into local optimum, thus avoiding a fast convergence of the population. In artificial neural networks, for example, the perturbation of some critical elements of the network during training is a common practice that improves its efficiency for pattern recognition [7]. We hypothesize that a population evolves better solutions when the fitness of the standard genetic cycle is affected by the social interaction between the members of the population. Our approach is inspired by the social life of microorganisms such as the social amoebas (*M. xanthus* and *D. discoideum*) [8, 9, 10] and viruses ($\phi$ 6) [11, 12, 13]. These social microorganisms, in response to specific environmental signals (starvation, reproduction, etc.), organize into social systems with two kinds of individuals: cooperators and defectors. Cooperators perform a group beneficial function whereas defectors are 'social parasites' or cheaters that fail to perform a group beneficial function. They instead reap the benefits of belonging to a group. Usually, in Nature cheaters are genetic mutants or individuals including selfish genes.

Based on above considerations we developed a GA with a population composed by cooperators and defectors individuals. In this approach the fitness function includes two terms:

$$f_i = f(x) + \Delta f(x) \quad (1)$$

The first term, *f(x)*, term is the standard chromosome fitness calculated by the objective function, whereas the second term, $\Delta f(x)$, is the fitness given by the social interaction between chromosomes. The $\Delta f(x)$ value is obtained applying a Game Theory model. Thus, it is the value given by the payoff matrix that summarizes the combination of strategies (cooperator, defector) during pair-wise interactions. We selected several games from general GT (see http://www.gametheory.net/), and GT applied to Microbiology [8, 13], namely: (i) prisoner's dilemma, (ii) chicken game, (iii) mixed polymorphism, (iv) friend or foe, (v) facultative defection (vi) battle of sexes, and (vii) stag hunt. All these games are: (a) 2x2, (b) symmetric and (c) non-zero sum games.

We are interested in identifying the game model that confers a population better performance than a standard GA. We conducted extensive experiments on three Knapsack instances (with single and multiple sacks). Our results show, for some game models, a significant improvement of the optimized solutions as compared to those obtained by a standard GA.

The next section outlines the proposed GA with social interactions. Thereafter, Section 3 describes in detail the simulation experiments conducted on both the single and multidimensional Knapsack problem. Section 4 shows our simulations results, Section 5 discusses the results, whilst section 6 summarizes and concludes our findings.

## 2. A GENETIC ALGORITHM WITH SOCIAL INTERACTIONS BETWEEN INDIVIDUALS

This section describes the proposed GA including a social interaction step.

### 2.1 A GAGT algorithm
The outline of the algorithm is the following:

/* GAGT Algorithm */
    WHILE not stop condition DO
      Social interaction.
      Evaluation ➔ $f_i$.
      Selection (or reproduction) of a new generation.
      Crossover.
      Mutation.
    END DO

### 2.2 Social interaction between individuals
Our approach assumes a mixed population composed by cooperators and defectors. Chromosome interactions are pair-wise, they interact equally likely with each other. The interactions are modeled according to GT. The GA is hybridized with the GT model as follows. Let *i* be a chromosome, its fitness value, $f_{i,}$, is given by two terms:

$$f_i = \beta_{GA} \frac{f(x)}{f^{max}} + \beta_{GT} \frac{\Delta f(x)}{\Delta f^{max}} \quad (2)$$

where *f(x)* is the 'genetic term' or fitness value calculated with the problem objective function, and $\Delta f(x)$ is a 'social term', which corresponds to the resulting payoff after the social interactions between chromosomes. In the above expression, both fitness terms are normalized. The normalization terms are the maximum fitness of the population at any given generation $f^{max}$ and the maximum payoff in the payoff matrix $\Delta f^{max}$. Note how the fitness value $f_i$ is a weighted sum of fitness values. In particular, $\beta_{GA}$ and $\beta_{GT}$ are the weights modeling the relevance of the genetic and social events, respectively.

As already mentioned, we assume a mixed population composed by two kinds of chromosomes: cooperators and defectors. The cooperators correspond to the usual GA chromosomes, whose fitness is calculated in conformity with regular practice in genetic algorithms. However, defector chromosomes exhibit a distinctive feature: they are able to cheat ('act dishonestly') when the fitness value is calculated. A defector will increase 'dishonestly' its fitness value compared with a cooperator. It is important to note that even when fitness is modified it does not involve a change in the chromosome gene values.

Let us consider a social interaction between two chromosomes selected at random. The combination of strategies (C=cooperator, D=defector) during a pair-wise interaction is summarized by the following payoff matrix:

$$\begin{array}{c|cc} & C & D \\ C & k & k-s_1 \\ D & k+s_2 & s-c \end{array} \quad (3)$$

which shows the payoff for the row player. In the matrix, *k* is the reward and $k-s_1$ the sucker's payoff that will be included in the fitness function of a cooperative chromosome. Likewise, in a cheater chromosome the following values, $k+s_2$ or temptation to cheat, and $k-c$ or punishment, will be also included in the fitness function. Considering the above payoff matrix we have four possible cases:

(a) A cooperative chromosome *i* meets with another cooperative chromosome, then the fitness value for the cooperative chromosome *i* is calculated with $\Delta f(x) =$ k.

(b) A cooperative chromosome *i* meets with a cheater chromosome, then the fitness value for the cooperative chromosome *i* is calculated with $\Delta f(x) =$ k-$s_1$.

(c) A cheater chromosome *i* meets with a cooperating chromosome, then the fitness value for the cheater chromosome *i* is calculated with $\Delta f(x) = k+s_2$.

(d) A cheater chromosome *i* meets with another cheater chromosome, then the fitness value for the cheater chromosome *i* is calculated with $\Delta f(x) = k-c$.

We conducted several simulation experiments with different social interaction models: prisoner's dilemma (PD), chicken game (CG), mixed polymorphism (MP), friend or foe (FOF), facultative defection (FD), battle of sexes (BS) and stag hunt (SH). The payoff matrix (3) could be replaced by this other equivalent matrix:

$$\begin{array}{cc} & C \quad D \\ C & R \quad S \\ D & T \quad P \end{array} \quad (4)$$

being R, S, T and P the reward, sucker's payoff, temptation to cheat, and punishment, respectively. In agreement with GT for each one of the social interaction models the payoffs will be equal to:

PD: T>R>P>S or $k+s_2>k>k-c>k-s_1$ with $c<s_1$

CG: T>R>S>P or $k+s_2 > k > k-s_1 > k-c$ with $c>s_1$

MP: T>R>S>P=0 or $k+s_2>k>k-s_1>k=c$ with $k=c>s_1$

FOF: T>R>P=S=0 or $k+s2>k> (k=c) = (k=s_1)$ with $k=c=s_1$

FD: T>R=P>S or $k+s_2>(k=k-c)>k-s_1$

BS: R>P>T=S=0 or $k>k-c>(k+s_2)=(k-s_1)=0$

SH: R>T>=P>S or $k>k+s_2>=k+s_2>k-s_1$

## 2.3 Genetic operators
The remaining steps of our algorithm closely resemble those of a standard GA. Specifically we used binary tournament selection, two-point recombination with a rate of 0.75, and the standard bit mutation with a rate of $1/L$, where $L$ is the length of the chromosome (in this case the number of items in the underlying Knapsack instance).

# 3. SIMULATION EXPERIMENTS

## 3.1 Single 0/1 Knapsack problem
Let *i* be a chromosome and assume we have *j* objects to be packed in a single sack. Each item has a value $v_j$ and weight $w_j$. With $W$ being the maximum weight that we can carry in the knapsack. We used the well-known 0-1-knapsack problem, restricting the number of each object $x_j$ to 0 or 1. The aim is to maximize $\sum_j v_j x_j$ subjected to $\sum_j w_j x_j \leq W$. We consider the chromosomes encoded as binary strings: a value of 1 indicates that an object is placed in the knapsack, whilst a value of 0 indicates that the object is left behind. The population is composed by two kinds of chromosomes: cooperative and cheater chromosomes. As we mentioned before, the difference between them lies is the way both genetic and social fitness are calculated. A cheater chromosome *i* will increases the value $v_j$ of an object *j* in an amount equal to $\Delta v_j$, or alternatively decreases its weight value $w_j$ to $\Delta w_j$. The fitness of both types of chromosomes will be calculated as follows. The fitness $g(x)$ for a cooperative chromosome is given by the usual expressions:

$$g(x) = \begin{cases} \sum_j w_j x_j \leq W, \sum_j v_j x_j \\ \sum_j w_j x_j > W, W - \sum_j w_j x_j \end{cases} \quad (5)$$

whereas for a cheater chromosome the fitness is given by:

$$g(x) = \begin{cases} \sum_j w_j x_j \leq W, \sum_j (v_j + \Delta v_j) x_j \\ \sum_j w_j x_j > W, W - \sum_j (w_j - \Delta w_j) x_j \end{cases} \quad (6)$$

such that $\Delta v_j = \tau/100$ and $\Delta w_j = \tau/100$ being $\tau$ the 'cheating degree' (i.e. 10, 20, …, 100).

Once two chromosomes are selected at random, a social interaction takes place between them. Three are the possible chromosome-chromosome interactions:

- Cooperative-cooperative:

$$f_i = \begin{cases} \sum_j w_j x_j \leq W, \beta_{GA} \dfrac{\sum_j v_j x_j}{f^{max}} + \beta_{GT} \dfrac{k}{\Delta f^{max}} \\ \sum_j w_j x_j > W, \beta_{GA} \dfrac{W - \sum_j w_j x_j}{f^{max}} + \beta_{GT} \dfrac{k}{\Delta f^{max}} \end{cases} \quad (7)$$

- Cooperative-cheater:

  Cooperative:

$$f_i = \begin{cases} \sum_j w_j x_j \leq W, \beta_{GA} \dfrac{\sum_j v_j x_j}{f^{max}} + \beta_{GT} \dfrac{k-s_1}{\Delta f^{max}} \\ \sum_j w_j x_j > W, \beta_{GA} \dfrac{W - \sum_j w_j x_j}{f^{max}} + \beta_{GT} \dfrac{k-s_1}{\Delta f^{max}} \end{cases} \quad (8)$$

Cheater:

$$f_i = \begin{cases} \sum_j w_j x_j \leq W \,,\, \beta_{GA}\dfrac{\sum_j (v_j+\Delta v_j)x_j}{f^{\max}}+\beta_{GT}\dfrac{k+s_2}{\Delta f^{\max}} \\ \\ \sum_j w_j x_j > W \,,\, \beta_{GA}\dfrac{W-\sum_j (w_j-\Delta w_j)x_j}{f^{\max}}+\beta_{GT}\dfrac{k+s_2}{\Delta f^{\max}} \end{cases} \quad (9)$$

- Cheater-cheater:

$$f_i = \begin{cases} \sum_j w_j x_j \leq W \,,\, \beta_{GA}\dfrac{\sum_j (v_j+\Delta v_j)x_j}{f^{\max}}+\beta_{GT}\dfrac{k-c}{\Delta f^{\max}} \\ \\ \sum_j w_j x_j > W \,,\, \beta_{GA}\dfrac{W-\sum_j (w_j-\Delta w_j)x_j}{f^{\max}}+\beta_{GT}\dfrac{k-c}{\Delta f^{\max}} \end{cases} \quad (10)$$

In the simulation experiments the conditions are given by (1) the optimization problem, i.e. knapsack problem, (2) the game theory parameters ($k$, $s_1$, $s_2$ and $c$ values of the payoff matrices should be scaled with the optimization problem) and (3) the genetic algorithm parameters (described in section 4).

Game model parameters, $k$, $s_1$, $s_2$ and $c$ values in the payoff matrices, were set up according to the values shown in Table 1.

**Table 1.- Game models. Payoff matrix parameters**[*]

|       | PD (1) | CG (2) | MP (1) | FOF (3) | FD (4) | BS (5) | SH (5) |
|-------|--------|--------|--------|---------|--------|--------|--------|
| $K$   | 1.0    | 1.0    | 1.0    | 1.0     | 1.0    | 1.0    | 1.0    |
| $S_1$ | 0.4    | 0.5    | 0.5    | 1.0     | 0.3    | 1.0    | 0.7    |
| $S_2$ | 1.0    | 9.0    | 0.5    | 0.5     | 0.3    | -1.0   | -0.2   |
| $c$   | 0.2    | 0.17   | 1.0    | 1.0     | 0.0    | 0.3    | 1.0    |

[*] Payoffs estimation from:

(1) see virus $\phi$ 6 [11, 12]. (2) see social amoeba *M. xanthus* [9, 10]. (3) Values were estimated based on (1). (4) see social amoeba *D. discoideum* [10]. (5) Usual values were normalized to 1.

In addition, we studied the relationship between the cheating degree $\tau$ and the number of feasible solutions obtained ($Y_{NS}$). The experiments were conducted with the PD model, $\beta_{GA}$=0.8, $\beta_{GT}$=0.2, and the following cheating degrees 10, 15, 20, 25, 30, 40 and 50.

## 3.2 Multidimensional Knapsack Problem

A multidimensional version of the problem consists of $m$ knapsacks of weights $W_1$, $W_2$, …, $W_m$ and $j$ objects with values $v_1$, $v_2$, …, $v_j$. The objective is to find a solution that guarantees that no knapsack is overfilled $\sum_j w_{jm} x_j \leq W_m$ and that yields maximum value $\sum_j v_{jm} x_j$. The fitness $g(x)$ for a cooperative chromosome is given by the usual expressions:

$$g(x) = \begin{cases} \sum_j w_{jm} x_j \leq W_m \,,\, \sum_j v_{jm} x_j \\ \\ \sum_j w_{jm} x_j > W_m \,,\, 0 \end{cases} \quad (11)$$

discarding infeasible solutions. Thus, for those solutions that violates the constraints the fitness is 0. In the case of a cheater chromosome the fitness is given by:

$$g(x) = \begin{cases} \sum_j w_{jm} x_j \leq W_m \,,\, \sum_j (v_{jm}+\Delta v_{jm}) x_j \\ \\ \sum_j w_{jm} x_j > W_m \,,\, W_m - \sum_j (w_{jm}-\Delta w_{jm}) x_j \end{cases} \quad (12)$$

## 4. RESULTS

We conducted experiments on 3 benchmark instances of the Knapsack problem, as described in Table 2.

**Table 2.- Knapsack problem instances**

| Instance name | No. of items | No. of sacks | Best known | Source |
|---------------|--------------|--------------|------------|--------|
| SK250         | 250          | 1            | Na         | [14]   |
| MK250         | 250          | 5            | 59312      | [4]    |
| MKSento1      | 60           | 30           | 7772       | [15]   |

The genetic algorithm parameters: $N$ (population size), $G$ (number of generations), $\alpha$ (cheater rate), pc (crossover rate), $p_m$ (mutation rate), and $\tau$ (cheating degree), were set as follows: $N$ =500, G=1000, $\alpha$ =0.1. In order to facilitate the search for improved solutions, the experiments were conducted with a high value of the cheating degree $\tau$ =50, (single knapsack) and $\tau$ =60 (multiple knapsack). The crossover rate $p_c$ and mutation rate per bit $p_m$ were 0.75 and 1.0/$L$ (where $L$ is the chromosome length) respectively. A control experiment was carried out with a standard genetic algorithm. The number of replicas (algorithm runs) was set to 100 for each game model. For the standard GA, a larger number of replicas was conducted, specifically, 100 x the number of game models explored. We consider an 'experiment', a set of runs according to the description above, including the standard GA and the seven game models studied.

Our results support that the inclusion of 'social interactions' modeled via Game Theory into the GA cycle improves the algorithm's performance. Figure 1 illustrates the best feasible solutions obtained for the single knapsack problem. For each solution, the plot show the knapsack weight (Y axis), and the knapsack value or fitness (X axis). We observe that the standard GA shows worse performance (left side, Figure 1) than those solutions obtained with the GA including social interactions (right side, Figure 1). It is important to note that whereas PD, FOF and FD promote cheating, BS and SH promote cooperation. CG and MP promote a mix of cheaters and cooperators (polymorphic populations). With CG the population evolves to a soft polymorphism, whilst with MP the population will exhibit a strong polymorphism (about 50%-50% of cheaters-cooperators).

Figure 2 shows a perfect linear regression (Table 3) between the cheating degree $\tau$ and the number of feasible solutions ($Y_{NS}$). We also found a perfect linear regression (Tables 4-5) between the cheating degree $\tau$ and the mean fitness of feasible solutions ($\overline{f_{NS}}$, Figure 3). A similar linear relationship is observed between $\tau$ and the maximum fitness value of feasible solutions ($f_{NS}^{max}$, Figure 4).

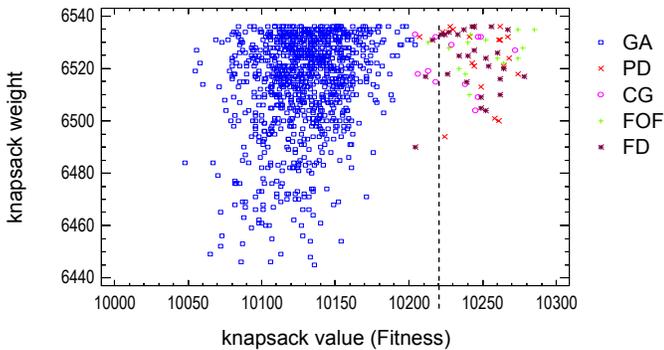

**Figure 1.** Single Knapsack problem. Best solutions founds with the standard GA (left) and the GA with social interactions (right).

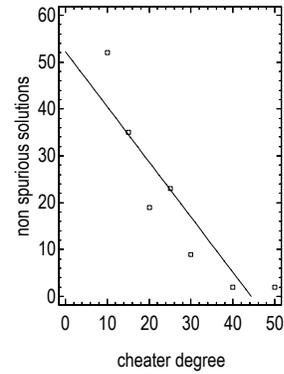

**Figure 2.** Linear regression between the cheating degree $\tau$ and the number of feasible solutions ($Y_{NS}$).

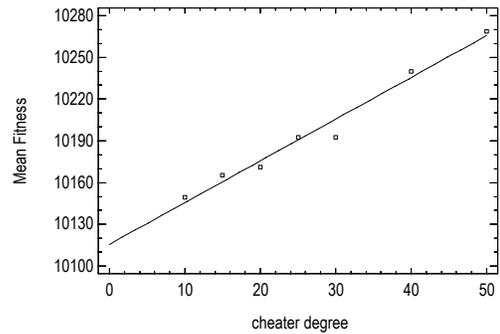

**Figure 3.** Linear regression between the cheating degree $\tau$ and the mean fitness of non-spurious solutions ($\overline{f_{NS}}$).

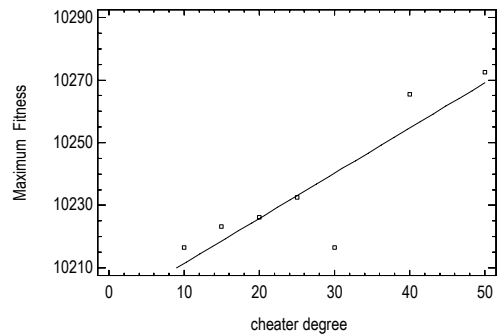

**Figure 4.** Linear regression between the cheating degree $\tau$ and the maximum fitness value of non-spurious solutions ($f_{NS}^{max}$).

**Table 3.- Regression ANOVA** ($Y_{NS}$ = 52.2395 - 1.17725 $\tau$ )

| Source | SS | DF | MSS | F-rate | P-Value |
|---|---|---|---|---|---|
| Model | 1653.19 | 1 | 1653.19 | 22.09 | 0.0053 |
| Residual | 374.24 | 5 | 74.8479 | | |

Total (Corr.) 2027.43    6

Correlation coefficient = - 0.903002

R-square = 81.5412

**Table 4.- Regression ANOVA** ( $\overline{f_{NS}}$ = 10115.6 + 3.0 $\tau$ )

| Source | SS | DF | MSS | F-rate | P-Value |
|---|---|---|---|---|---|
| Model | 10735.7 | 1 | 10735.7 | 204.97 | 0.0000 |
| Residual | 251.88 | 5 | 52.376 | | |

Total (Corr.) 10997.6    6

Correlation coefficient = 0.988022

R-square = 97.6188

**Table 5.- Regression ANOVA** ( $f_{NS}^{\max}$ = 10197.0 + 1.44263 $\tau$ )

| Source | SS | DF | MSS | F-rate | P-Value |
|---|---|---|---|---|---|
| Model | 2482.57 | 1 | 2482.57 | 16.77 | 0.0094 |
| Residual | 740.089 | 5 | 148.018 | | |

Total (Corr.) 3222.66    6

Correlation coefficient = 0.877695

R-square = 77.0348

Regarding the Multiple Knapsack instances, for *Sento1* we obtained, in two different experiments with $\tau$ =50, a total of 10 solutions significantly better than those obtained with the standard GA. For the increased cheating degree to $\tau$ =60, 6 solutions were significantly better than those obtained in the control experiment. In the three experiments carried out with this instance the best results were obtained with FOF, FD and PD. The CG game failed to produce better results than those obtained in the control experiments. In the two experiments conducted under $\tau$ =50, the best solution 7719 was found with the FOF and FD models. A Mann-Whitney (Wilcoxon)'s test (with a *p*-value equal to zero) show that the differences among medians were statistically significant at the 95.0% confidence level. In consequence, the GA with social interactions performs better than the standard GA regardless of the game model used. In the experiment with $\tau$ =60 the Mann-Whitney (Wilcoxon) test, with a *p*-value 0.013, show that there are statistically significant differences among the medians at the 95.0% confidence level. Once again, we conclude that our approach improves the algorithm's performance.

Regarding the multiple knapsack instance *MK250*, the social GA produced good results in only two of the three experiments. In these two experiments a total of 8 significant best, as compared to the best result in the control experiment, were obtained. The best results were obtained with the FD game model, being the best obtained solution value 56481. The Mann-Whitney (Wilcoxon) test (with a *p*-value 0.001) show that there are statistically significant differences among the medians at the 95.0% confidence level.

In order to strengthen or support our results we performed some experiments replacing the social interactions (payoff matrix values) by uniform or Gaussian noise (stochastic matrix). The experiments were conducted with PD model, $\beta_{GA}$=0.8, $\beta_{GT}$=0.2, and *SK100* [14]. Figure 5 demonstrates how solutions obtained in presence of noise are well below of those obtained including social interactions into the GA cycle.

## 5. DISCUSSION

Our interpretation of the dynamic behavior of a genetic algorithm with social interactions is as follows. Initially, cooperators and only a few cheater chromosomes compose the population, starting out the evolution of such mixed-population. After many generations, a solution with high fitness is reached by a cheater chromosome. At this stage most solutions, cooperators and cheaters, are non feasible (spurious) solutions (e.g. knapsacks with high $\sum_{j} v_j x_j$ but $\sum_{j} w_j x_j > W$ ) being the population on the verge of extinction. Only a few cheater chromosomes are optimized feasible solutions, showing a significant improvement compared with a simple GA. When solving a optimization problem, the interest lies in finding particularly good solutions, and the final fate of population is meaningless [16]. It is important to note that our approach promotes cheaters whereas the usual approach in Game Theory or in Evolutionary Game Theory promotes cooperation. For instance, some mechanisms have been suggested to promote cooperation such as chaotic variations of appropriate amplitude [17], kin selection [18], tit-for-tat strategy [19], etc.

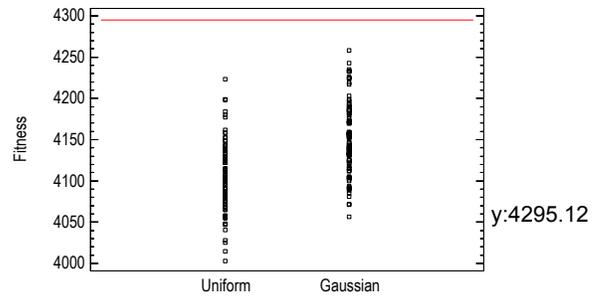

**Figure 5. Single Knapsack problem with uniform or Gaussian noise. Above the threshold line (4295) are the solutions obtained with social interactions (PD model).**

In our social GA, however, cheaters 'promote themselves' by having an eventual reproductive (fitness) advantage. In consequence, no additional mechanism is needed to promote cheating. We propose that modeling cheating through a genetic algorithm with social interactions could be a novel approach for problem solving.

## 6. SUMMARY AND CONCLUSIONS

We have proposed a variation of the standard GA that includes social interactions between the members of the population. The social interaction is modeled according to Game theory, and a number of well known game models in the theoretical biology literature were studied. We found that the proposed social interaction step improves the problem solving capabilities of the standard genetic cycle. Therefore, we suggest that modeling cheating through a genetic algorithm with social interactions could be a novel approach for problem solving.

Our future work will include a larger number of instances and additional combinatorial problems. Moreover, the implementation of the proposed technique in the context of Genetic Programming is currently under study.

## ACKNOWLEDGMENTS
The first author was supported by Laboratorio de Bioinformatica, Complutense University of Madrid (UCM), and in part under 'Profesores UCM en el Extranjero 2008' grant.